\documentclass[letterpaper]{article} 
\usepackage{aaai23}  
\usepackage{times}  
\usepackage{helvet}  
\usepackage{courier}  
\usepackage[hyphens]{url}  
\usepackage{graphicx} 
\usepackage{natbib}  
\usepackage{caption} 
\usepackage{algorithm}
\usepackage{algorithmic}
\usepackage{multirow}
\usepackage{amsmath}
\usepackage{booktabs}
\usepackage{utfsym}
\usepackage{pgfplots}
\usepackage{amsfonts}

\usepackage{newfloat}
\usepackage{listings}
\DeclareCaptionStyle{ruled}{labelfont=normalfont,labelsep=colon,strut=off} 
\floatstyle{ruled}
\newfloat{listing}{tb}{lst}{}
\floatname{listing}{Listing}
\title{Knowledge-Bridged Causal Interaction Network for Causal Emotion Entailment}
\author{
    Weixiang Zhao,
    Yanyan Zhao \thanks{Corresponding Author},
    Zhuojun Li,
    Bing Qin
}
\affiliations{
    Research Center for Social Computing and Information Retrieval\\

    Harbin Insititute of Technology, China\\
    \{wxzhao, yyzhao, zjli, qinb\}@ir.hit.edu.cn
}

\usepackage{bibentry}

\begin{document}

\maketitle

\begin{abstract}
Causal Emotion Entailment aims to identify causal utterances that are responsible for the target utterance with a non-neutral emotion in conversations. Previous works are limited in thorough understanding of the conversational context and accurate reasoning of the emotion cause. To this end, we propose Knowledge-Bridged Causal Interaction Network (KBCIN) with commonsense knowledge (CSK) leveraged as three bridges. Specifically, we construct a conversational graph for each conversation and leverage the event-centered CSK as the semantics-level bridge (S-bridge) to capture the deep inter-utterance dependencies in the conversational context via the CSK-Enhanced Graph Attention module. Moreover, social-interaction CSK serves as emotion-level bridge (E-bridge) and action-level bridge (A-bridge) to connect candidate utterances with the target one, which provides explicit causal clues for the Emotional Interaction module and Actional Interaction module to reason the target emotion. Experimental results show that our model achieves better performance over most baseline models. Our source code is publicly available at \url{https://github.com/circle-hit/KBCIN}.

\end{abstract}

\section{Introduction}
Emotion analysis in conversations has become an emerging topic in natural language processing (NLP) community. Most existing works mainly focus on Emotion Recognition in Conversations (ERC), which aims at predicting the emotion label for each utterance in conversations \cite{dialoguernn,cosmic,dag-erc}. However, emotion-reasoning task such as recognizing the cause behind emotions in conversations is yet under-explored. More recently, Poria et.al \shortcite{reccon} holds that Recognizing Emotion Cause in CONversations (RECCON) is beneficial to improve interpretability and performance in affect-based models. Also, it has potential applications in several areas such as emotional support system \cite{es_peng} and empathetic dialog system \cite{empathy1}. Thus, Poria et.al \shortcite{reccon} introduce a new task named RECCON with an annotated dataset. It includes two different sub-tasks: Causal Span Extraction (CSE) and Causal Emotion Entailment (CEE). 
\begin{figure}[htbp]
\centering
\includegraphics[width=0.5\textwidth]{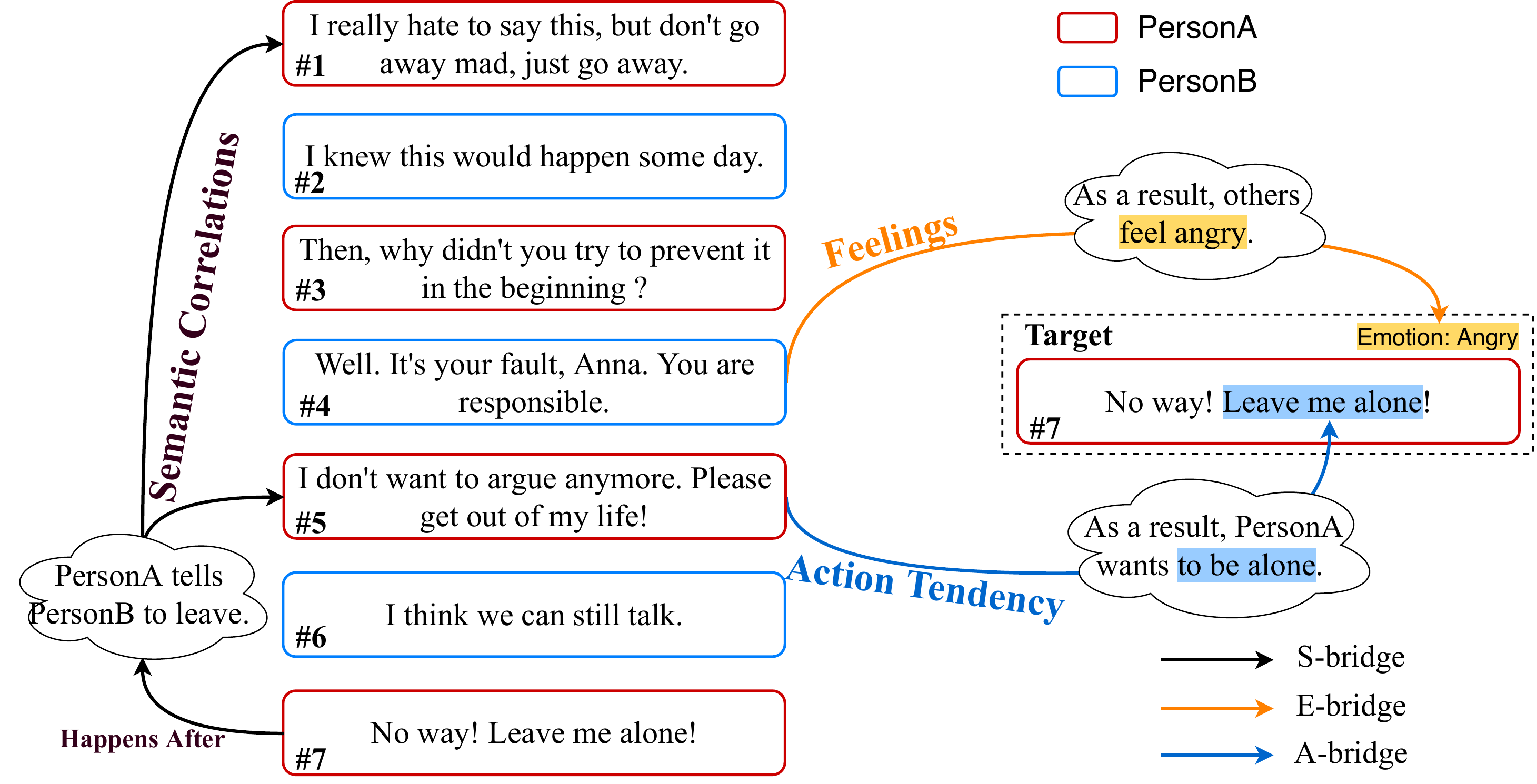}
\caption{An example from RECCON-DD dataset \cite{reccon} for identifying the emotion cause with the help of S-bridge, E-bridge and A-bridge.}
\label{example}
\end{figure}
We focus on the CEE sub-task in this paper and its goal is to predict which particular utterances in the conversation history trigger the non-neutral emotion in the target utterance.

There are two main challenges in the CEE task. First, to cope with the intermingling emotional dynamics among interlocutors, it is necessary to comprehend the deep semantic correlations between contextual utterances via effective conversational context modeling. Second, it may be difficult to accurately reason candidate utterances to the target emotion because causal clues are not always explicitly mentioned in the context but are supposed to be implied through reasoning based inference, which results in a reasoning gap between candidates and the target. However, Poria et.al \shortcite{reccon} simply formulate CEE as an utterance-pair classification problem, which is lack of sufficient conversational context modeling and effective emotion cause reasoning. Thus, to deal with such two challenges, we introduce commonsense knowledge (CSK) \cite{csk} into CEE.


\textbf{On the one hand}, the event-centered CSK, which manifests what may happen before or after the event mentioned in an utterance, could be viewed as the semantics-level bridge (S-bridge) to connect the development of a conversation and enhance the semantic dependencies between relevant utterances, leading to the thorough understanding of the conversational context. As shown in the left of Figure \ref{example}, the event in utterance $\#$7 that PersonA wants to leave alone happens after the event \emph{PersonA tells PersonB to leave}, which is associated with utterances $\#$1 and $\#$5.

\textbf{On the other hand}, according to \cite{theory}, feeling and action tendency of human beings are two important components of emotion and contribute largely to offer potential causal clues for the generation of target emotions. To this end, social-interaction CSK is leveraged as the emotion-level bridge (E-bridge) and the action-level bridge (A-bridge) to connect candidate utterances with the target one according to causal clues conveyed by the feeling and action tendency from interlocutors. In Figure \ref{example}, the excuse and criticism from PersonB in utterance $\#$4 make PersonA \emph{feel angry}, which is consistent with the emotion carried by the target utterance $\#$7. Further, the content of utterance $\#$5 implies the action tendency of PersonA \emph{to be alone} and it directly causes what she expresses in the target utterance $\#$7.

In this paper, we propose Knowledge-Bridged Causal Interaction Network (KBCIN) to effectively carry out conversational context modeling and emotion cause reasoning. Specifically, we abstract a conversation as a conversational graph to model the inter-utterance dependencies in the conversation. Then event-centered CSKs including \emph{isAfter} and \emph{isBefore} are introduced and we devise CSK-Enhanced Graph Attention module to integrate CSKs as S-bridge for message passing on the graph. Further, to fill the reasoning gap between candidate utterances and the target one, social-interaction CSKs \emph{x(o)Want} and \emph{x(o)React} are leveraged as A-bridge and E-bridge. We design Emotional Interaction module and Actional Interaction module to accurately reason the cause of the target emotion with the help of the explicit causal clues conveyed by the two bridges. And the above three modules form the Knowledge-Bridged Causal Interaction (KBCI) block and it is paralleled as multiple heads to sufficiently model the inter-dependencies among utterances and precisely associate the target emotion with candidate utterances. 

To evaluate the performance of the proposed model, we conduct extensive experiments on the benchmark dataset \cite{reccon}. State-of-the-art performance is achieved by us compared with the baseline models of CEE and other strong baslines on the task of Emotion Cause Extraction (ECE) and Emotion-Cause Pair Extraction (ECPE).

The main contributions of this work are summarized as follows:
\begin{itemize}
\item We introduce commonsense knowledge into the Causal Emotion Entailment task to fill the reasoning gap between candidate utterances and the target one.
\item We propose a novel model KBCIN to perform comprehensive conversational context modeling and accurate emotion cause reasoning with commonsense knowledge as three bridges.
\item Extensive experimental results over most strong baselines on the benchmark dataset demonstrate the superiority of our model.
\end{itemize}

\begin{figure}[htbp]
\centering
\includegraphics[width=0.35\textwidth]{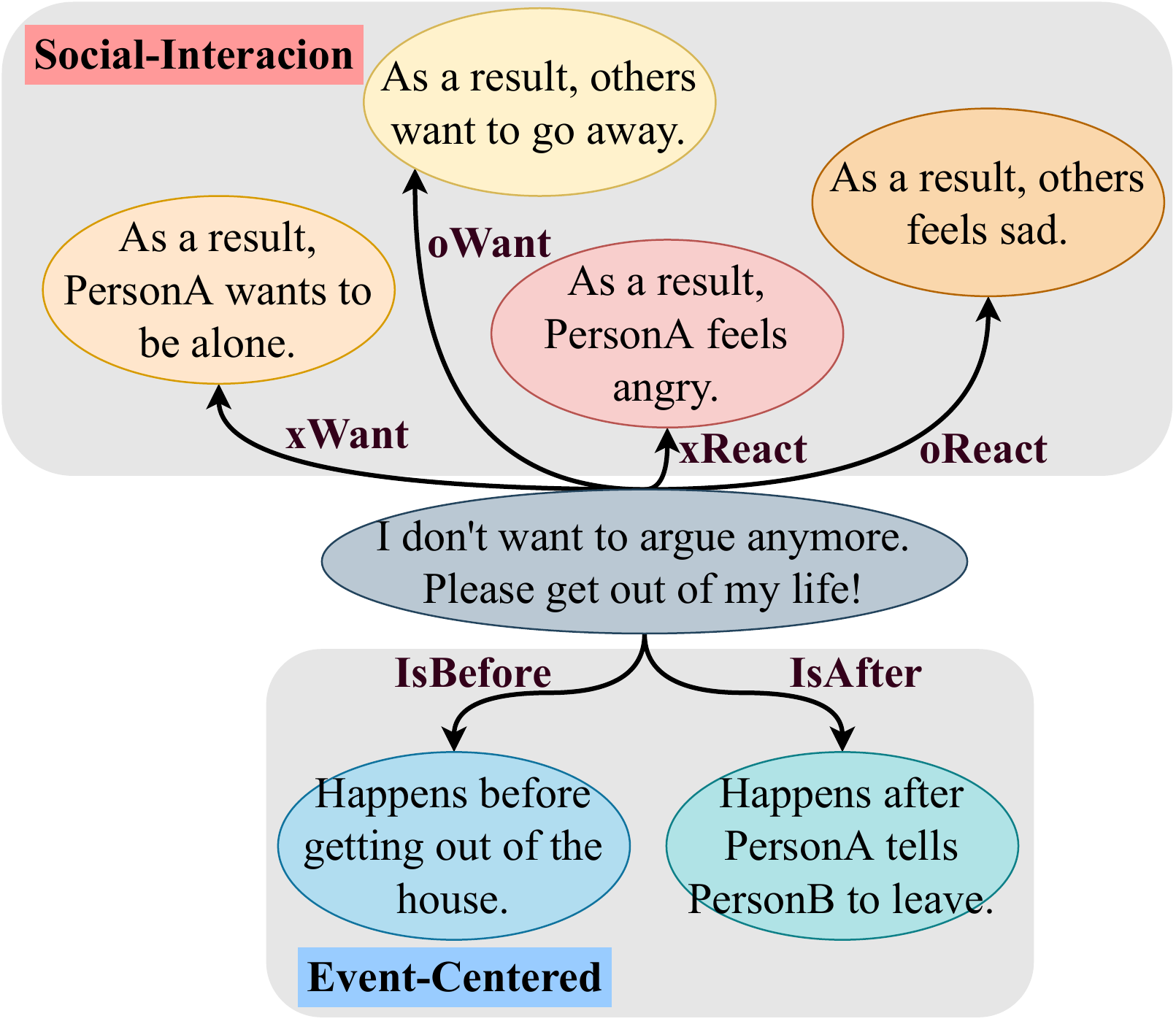}
\caption{Examples of the event-centered and social-interaction CSK.}
\label{csk_example}
\end{figure}

\section{Related Work}

\subsection{Causal Emotion Entailment}
Poria et.al \shortcite{reccon} propose the task of recognizing emotion cause in conversations and define two novel sub-tasks named Causal Span Extraction (CSE) and Casual Emotion Entailment (CEE) to identify the emotion cause in the span-level and utterance-level, respectively. They formulate CEE as an utterance-pair classification task and build strong transformer-based baselines. Based on the emotion recognition model proposed by Shen et.al \shortcite{dag-erc}, Li et.al \shortcite{kec} construct knowledge-enhanced conversation graph and propagate causal clues through it.

Targeting at the two main challenges, we leverage CSK as three bridges to perform effective conversational context modeling and fill the reasoning gap between candidate utterances and the target to accurately identify the emotion cause.

\subsection{Emotion Cause Extraction}
Apart from CEE, there are other two emotion cause related tasks named Emotion Cause Extraction (ECE) and Emotion Cause Pair Extraction (ECPE).

Gui et.al \shortcite{ece-data} construct the typical ECE dataset and the goal of this task is to extract cause clauses which express triggers leading to the target emotion expressed in the emotion clause. To effectively identify cause clauses, a popular strategy is to leverage relative position information to associate candidate clauses with the emotion clause \cite{ding2019,xia2019,li2019}. However, these works are proved to suffer from the problem of position bias and may not generalise well. Thus, recent works on ECE attempt to mitigate the position bias problem by enhance deeper semantic dependencies between the emotion clause and cause clauses \cite{kag}.

Xia and Ding \shortcite{ecpe} propose the ECPE task, with the goal to extract the potential pairs of emotions and corresponding causes in the document. Compared to ECE task, ECPE is more closer to practical applications. Previous works project the representations of every clause into a matrix to form candidate pairs and assign a confidence score to each pair. According to different ways of training, they could be categorized as two-stage method \cite{ecpe-2d} and end-to-end method \cite{ecpe-mll,rankcp,ece-2022}.

Different from these two tasks which extracts emotion cause in news articles, CEE focuses on the conversation scenario, which is particularly challenging due to the complex emotional interactions.

\begin{figure*}[htbp]
\centering
\includegraphics[width=0.7\textwidth]{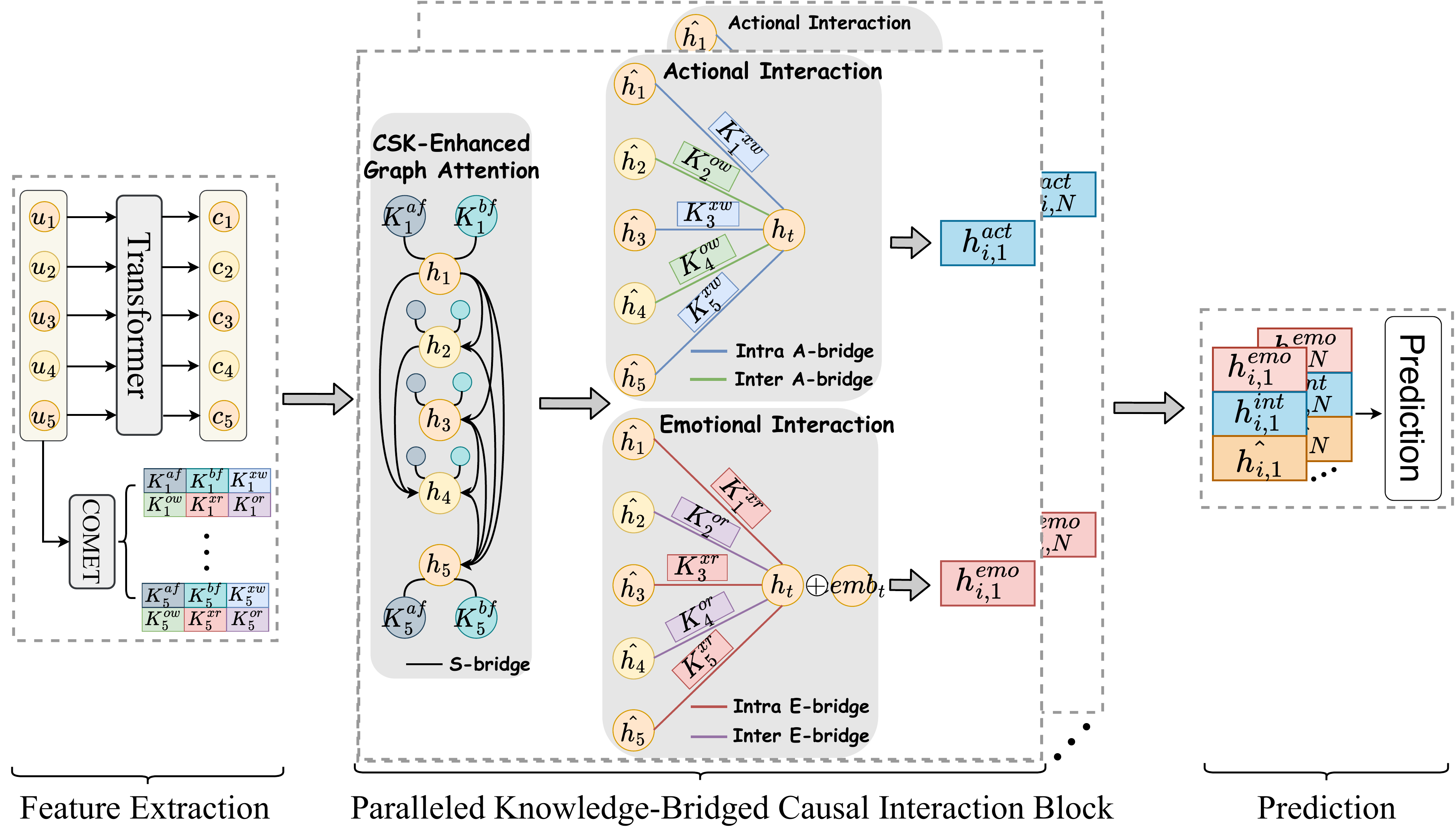}
\caption{The overall architecture of our proposed model.}
\label{model}
\end{figure*}

\section{Methodology}
\subsection{Task Definition}
First, we define the problem of the CEE task. Given a conversation that consists of $t$ consecutive utterances $\left\{u_1, u_2, \cdots, u_t \right\}$ with the corresponding emotion label $\left\{e_1, e_2, \cdots, e_t \right\}$ between two speakers, the goal of this task is to predict which particular utterances $u_i$ ($i \le t$) in the conversational history are responsible for the non-neutral emotion $e_t$ in the target utterance $u_t$. And $u_i$ is a positive example if it contains the cause of non-neutral emotion in the target utterance and a negative example otherwise. The architecture of our proposed model KBCIN is shown in Figure \ref{model}. We assume that $\left\{u_1, u_2, \cdots, u_5 \right\}$ is the input conversation and $u_5$ is the target utterance. Utterances from the same speaker are in the same color.

\subsection{Feature Extraction}
\paragraph{Utterance-Level Feature Extraction.}
Transformer encoder \citep{trans} is adopted as the utterance encoder to extract utterance-level features. Specifically, for each utterance $u_i = \left\{w_1, w_2, \cdots, w_L \right\}$, a special token $[CLS]$ is concatenated to the beginning of the utterance. Then we feed the sequence to the utterance encoder and the representation after max-pooling from the last hidden layer is obtained as the utterance-level feature of each utterance.
\begin{equation}
    c_i = \textrm{Maxpooling}(\textrm{Transformer}(\,[CLS], w_1, w_2, \cdots, w_L\,))
\end{equation}
where $c_i \in \mathbb{R}^{d_m}$ and $d_m$ is the dimension of hidden states in utterance encoder. Then each utterance vector $c_i$ is transformed to the dimension of $d_h$ with a linear projection.

\paragraph{Knowledge Acquisition.}
In this work, we use ATOMIC-2020 \cite{csk} as our commonsense knowledge (CSK) base. It is a CSK graph covering social, physical, and eventive aspects of everyday inferential knowledge.

To fully comprehend semantic dependencies among utterances in the conversation and fill the reasoning gap between candidate utterances and the target, we leverage CSK as three bridges, named semantics-level bridge (S-bridge), emotion-level bridge (E-bridge) and action-level bridge (A-bridge). To be more specific, we explore six types of CSK from ATOMIC-2020, which are categorized as event-centered CSK and social-interaction CSK. Examples of the CSK are shown in Figure \ref{csk_example}. On the one hand, the deep semantic dependencies among utterances would be built by associating more relevant utterances according to the temporality and causality of conversation development manifested by the event-centered CSK \emph{isAfter} and \emph{isBefore}. Thus, S-bridge is constructed to perform the comprehensive understanding of the conversational context. On the other hand, the other two bridges, E-bridge and A-bridge are built by social-interaction CSK \emph{xReact}, \emph{oReact}, \emph{xWant} and \emph{oWant}. And \emph{x(o)Want} is the description of what would PersonX (others) likely want to do after the event, while \emph{x(o)React} manifests the emotional feelings that how does PersonX (others) feel after the event. They serve to fill the reasoning gap between candidate utterances and the target one from the perspective of feeling and action tendency of human beings.

In order to generate representations of CSK for given utterances, we adopt the generative commonsense transformer model COMET \cite{comet}, which is trained on ATOMIC-2020. More specifically, we use the BART-based \cite{bart} variation of COMET. Given each utterance $u_i$ in a conversation to form the input format $(u_i, r, [\textrm{GEN}])$, where $r$ is the type of CSK we select, COMET would generate descriptions of inferential content under the relation $r$. And the hidden state representation from the last layer of COMET are adopted as the CSK representation. Through this, for each utterance $u_i$, six pieces of CSK representation are prepared for conversational context modeling and emotion cause reasoning. They are denoted as $K^r_i, r \in \left\{af, bf, xr, or, xw, ow\right\}$ and $af, bf, xr, or, xw, ow$ are short for the relation type \emph{isAfter}, \emph{isBefore}, \emph{xReact}, \emph{oReact}, \emph{xWant} and \emph{oWant}, respectively.

\subsection{Paralleled Knowledge-Bridged Causal Interaction}
Inspired by the notion of multi-head attention \cite{trans}, we propose paralleled Knowledge-Bridged Causal Interaction block which is devised to fully comprehend the conversational context and accurately reason the cause of the the non-neutral emotion in the target utterance. For each block, it consists of three components: CSK-enhanced graph attention module, emotional interaction module and actional interaction module.

\paragraph{CSK-Enhanced Graph Attention Module.}
Instead of formulating CEE as an utterance-pair classification problem without explicit modeling of utterance interactions, we abstract utterances in a conversation as a conversational graph where the current utterance only connects to the past utterances in the dialogue history. Through this, we make sure the interaction of utterances to meet the nature of causality that cause could only be reasoned from the past. And the representation of each node is initialised by the corresponding utterance-level feature $c_i$. In addition, we calculate the relative distance between the target utterance and candidate utterances and  utilize the relative position information to enrich the utterance representation. Since the emotion of each utterance is proved to play an important role in CEE \cite{reccon}, we also take it into consideration. Thus, the final representation of each node is obtained by: 
\begin{equation}
    h_i = c_i \oplus pemb_i \oplus eemb_i
\end{equation}
where $pemb_i$ represents the relative position embedding between $u_i$ and $u_t$ and $eemb_i$ is the emotion embedding of $e_i$.

Based on the vanilla graph attention network \cite{gat}, we devise the CSK-enhanced graph attention to propagate information on the conversational graph and leverage event-centered CSK as S-bridge to measure the semantic dependencies among utterances. The graph attention operated on the node representation to update it from the information of other neighbourhoods can be written as:
\begin{equation}
    \hat{h_i} = \sigma(\sum_{j \in N_i} \alpha_{ij} W_h h_j)
\end{equation}
where $N_i$ is the neighbors of node $i$, $W_h \in \mathbb{R}^{d_h \times d_h}$ is the trainable weight matrix and $\sigma$ represents the nonlinearity activation function.

The weight $\alpha_{ij}$ is utilised to measure the importance and relevance between the current node and its neighbours. We introduce event-centered CSK $K^{af}$ and $K^{bf}$ into the measuring process:
\begin{equation}
    \alpha_{ij} = \frac{\mathrm{exp}(\mathcal{F}(h_i, h_j, K^{af}_j, K^{bf}_j))}{\sum_{j^{\prime} \in N_i} \mathrm{exp}(\mathcal{F}(h_i, h_{j^{\prime}}, K^{af}_{j^{\prime}}, K^{bf}_{j^{\prime}}))}
\end{equation}
where $\mathcal{F}$ is an attention function.

Different from the attention function that purely calculates attention score between utterance representations \cite{gat}, we leverage the event-centered CSK $K^{af}$ and $K^{bf}$ as the S-bridge to measure the utterance dependencies. 
\begin{equation}
\begin{aligned}
    \mathcal{F}(h_i, h_j, K^{af}_j, K^{bf}_j) &= \mathrm{LeakyReLU}(a^\top [W_h h_i \parallel \\ W_h h_j & + W_e K^{af}_j + W_e K^{bf}_j])
    \end{aligned}
\end{equation}
where $a \in \mathbb{R}^{2d_h}$ and $W_e \in \mathbb{R^{d_h \times d_h}}$ are both trainable weight matrices.

\paragraph{Emotional Interaction Module.}
After the comprehensive conversational context modeling with the help of S-bridge, we leverage two types of social-interaction CSK, $K^{xr}$ and $K^{or}$ as the E-bridge to fill the reasoning gap and reason the target emotion according to the emotional causal clues. This idea is inspired by the theory that feelings are one of the most important components of human's emotion \cite{theory}. Thus, the target utterance with the corresponding emotion is more relevant to those candidates that could generate the most similar emotion or feeling as what the target holds. Further, to distinguish intra-speaker dependency and inter-speaker dependency, $K^{xr}$ and $K^{or}$ serve as intra E-bridge and inter E-bridge, respectively. The emotional similarity score can be obtained by:
\begin{equation}
\begin{aligned}
    Q^{emo} = f_q(h_t + eemb_t), \ K^{emo} = f_k(\hat{h_i}) + f_e(K^{r}_i) \\
    s_i^{emo} = \mathrm{softmax}(\frac{Q^{emo} (K^{emo})^\top}{\sqrt{d_h}})
    \end{aligned}
\end{equation}
where $f_q(x)$; $f_k(x)$; $f_e(x)$ are all linear transformations. $t$ is the index of the target utterance and $i \ (i \le t)$  is that of candidate utterance in the dialogue history. And $r=xr$ if the target utterance $u_t$ is of the same speaker with the candidate utterance $u_i$, otherwise $r=or$.

Then we utilize the emotional similarity score $s_i^{emo}$ to weight the importance of candidate utterances and enrich them with the representation of the target utterance:
\begin{equation}
\begin{aligned}
    h^{emo}_i = &s_i^{emo} V^{emo} + s_i^{emo} Q^{emo}\\
    V^{emo} = &f_v(\hat{h_i}) + f_e(K^{r}_i) \\
    \end{aligned}
\end{equation}
where $f_v(x)$ is the linear transformation.

\paragraph{Actional Interaction Module.}
Since action tendency is another important component to reason the aroused emotion of human beings, the other two types of social-interaction CSK $K^{xw}$ and $K^{ow}$ function as the A-bridge to make candidate utterances connected to the target one with the implied consistent action tendency. Also, intra A-bridge and inter A-bridge are formed. The actional similarity score is obtained by:
\begin{equation}
\begin{aligned}
    Q^{act} = f_q^{\prime}(h_t), \ K^{act} = f_k^{\prime}(\hat{h_i}) + f_e^{\prime}(K^{r}_i) \\
    s_i^{act} = \mathrm{softmax}(\frac{Q^{act} (K^{act})^\top}{\sqrt{d_h}})
    \end{aligned}
\end{equation}
where $f_q^{\prime}(x)$; $f_k^{\prime}(x)$; $f_e^{\prime}(x)$ are linear transformations. $r=xw$ if the speaker of the target utterance $u_t$ is same as that of the candidate utterance $u_i$, otherwise $r=ow$.

The weighed representation after actional interaction is:
\begin{equation}
\begin{aligned}
    h^{act}_i = &s_i^{act} V^{act} + s_i^{act} Q^{act}\\
    V^{act} = &f_v^{\prime}(\hat{h_i}) + f_e^{\prime}(K^{r}_i) \\
    \end{aligned}
\end{equation}
where $f_v^{\prime}(x)$ is the linear transformation.

Finally, at the end of each knowledge-bridged causal interaction block, to synthesize results in the reasoning process, we add the conversational representation $\hat{h_i}$, emotional representation $h_i^{emo}$ and actional representation $h_i^{act}$ together, and the final representation of each utterance is:
\begin{equation}
    \tilde{h_i} = \hat{h_i} + h_i^{emo} + h_i^{act}
\end{equation}

\subsection{Causal Utterance Prediction}
Here, taking the concatenation of the causal representations from each paralleled KBCI head as input, we utilise a causal utterance predictor to decide whether the candidate $u_i$ is the cause of the target $u_t$:
\begin{equation}
    \hat{y}_i = \mathrm{sigmoid}(\mathrm{MLP}(\parallel_{n=1}^N \tilde{h_{i,n}}))
\end{equation}
where $\parallel$ is the concatenation operation, MLP represents the multi-layer perception, and $N$ is the number of knowledge-bridged causal interaction block head.

\section{Experiments}
\subsection{Dataset and Evaluation Metrics}
We conduct experiments on the benchmark dataset RECCON-DD. It is collected from the popular dataset DailyDialog \cite{dd} with utterance-level emotion labels and the emotion cause labels are annotated by Poria et.al \shortcite{reccon}. We only take causes of the dialogue history into consideration and the repetitive causal pairs are removed. Statistics of the processed RECCON-DD are shown in Table \ref{tab1}.

Following Poria et al. \shortcite{reccon}, we report the F1 scores of both negative and positive causal pairs and the macro F1 scores of them.

\begin{table}
\footnotesize
\centering
\begin{tabular}{cccc}
\toprule
& \textbf{Train} & \textbf{Valid} & \textbf{Test} \\
\midrule
Positive Causal Pairs &7,027 &328 &1,767 \\
Negative Causal Pairs &20,646 &838 &5,330 \\
Num. of Dialogue &834 &47 &225 \\
Num. of Utterance &8,206 &493 &2,405\\
\bottomrule
\end{tabular}
\caption{Dataset statistics}
\label{tab1}
\end{table}

\subsection{Baselines and Comparison Models}
We compare our proposed model with baselines of CEE task. Since there are only few works on CEE, we also take baselines from other emotion cause related tasks, ECE and ECPE, into consideration.

{
\setlength{\parindent}{0cm}
\textbf{Methods for CEE:}
}

\textbf{RoBERTa-Base/Large} \cite{reccon} adopts the widely used pretrained language model from \cite{roberta}. In this setting, CEE is defined as a utterance-pair classification problem. They concatenate the special token [CLS], emotion label, target utterance, candidate utterance and the dialogue history to create the input.

\textbf{KEC} \cite{kec} extends the Directed Acyclic Graph networks (DAG) for ERC \cite{dag-erc} to build Knowledge Enhanced DAG networks. They leverage social CSK to boost the performance of identifying causal utterances with neutral emotion. It is worth to mention that neutral utterances are also involved as targeted utterances, leading to more negative non-causal pairs in their processed dataset. 

{
\setlength{\parindent}{0cm}
\textbf{Methods for ECE:}
}

\textbf{KAG} \cite{kag} proposes a novel graph-based method to alleviate the problem of position bias by leveraging the entity-related CSK to enhance the semantic dependencies between a candidate clause and an emotion clause.

\textbf{Adapted} \cite{adapted} jointly performs emotion and emotion cause recognition by combining CSK via adapted knowledge models with multi-task learning.

{
\setlength{\parindent}{0cm}
\textbf{Methods for ECPE:}
}

\textbf{ECPE-2D} \cite{ecpe-2d} proposes an end-to-end approach for emotion-cause pair extraction. The emotion-cause pairs are represented by a 2D representation scheme and 2D transformers are devised to model the interactions of different emotion-cause pairs.

\textbf{ECPE-MLL} \cite{ecpe-mll} proposes two joint frameworks for emotion-cause pair extraction: extraction of the cause clauses corresponding to the specified emotion clause and extraction of the emotion clauses corresponding to the specified cause clause.

\textbf{RankCP} \cite{rankcp} deals with the emotion-cause pair extraction by ranking clause pair candidates in a document and proposes a one-step neural approach to perform end-to-end extraction with inter-clause modeling.

\subsection{Implementation Details}
For utterance-level feature extraction, the dimension of hidden states in utterance encoder is 768, and the number of transformer encoder layer is 8 with 10 attention heads. Layers of emotion embedding and relative position embedding are randomly initialized and the dimension of both embedding layers are 300. Also, for all representations in the following parts of KBCIN, $d_h$ is set to 300. For causal utterance prediction, dimensions of MLP is set to [300, 300, 300, 1] and the dropout rate is set to 0.07. We utilize AdamW optimizer with learning rate of 4e-5 and L2 regularization of 3e-4 to train our model. And the batch size is 8. We pick the model which works best on the valid set, and then evaluate it on the test set. All of our results are averaged on 5 runs.

\section{Results and Analysis}
\subsection{Overall Results}
As shown in Table \ref{tab2}, our proposed model achieves state-of-the-art results on REECON-DD dataset. Since results of RoBERTa-Base/Large and methods for ECPE are achieved under the same dataset scale with ours, we directly refer them from Poria et.al \shortcite{reccon} and we reimplement KEC and methods for ECE in the same setting with us. Benefiting from the effective conversational context modeling through S-bridge and accurate emotion cause reasoning with E-bridge and A-bridge, KBCIN achieves state-of-the-art Pos. F1 and macro scores of 68.59 and 79.12, respectively.

For ECE and ECPE baselines, they are not comparable with our proposed KBCIN. It suggests that directly transferring the methods of context modeling and target measuring for article documents may not be suitable enough under the circumstance of conversation scenario. Also, the undesired results of KAG may be ascribed to the reason that entity-related CSK enhance the semantic dependencies among clauses to some extent, but it fails to offer valuable social-interaction causal clues to reason the target emotion.

\begin{table}
\footnotesize
\centering
\begin{tabular}{cccc}
\toprule
\textbf{Model} & Neg. F1 & Pos. F1 & macro F1\\
\midrule
\multicolumn{4}{c}{\textbf{ECE Methods}}  \\
KAG &86.35 &58.18 &72.26\\
Adapted  &88.18 &64.53 &76.36\\
\midrule
\multicolumn{4}{c}{\textbf{ECPE Methods}}  \\
ECPE-2D &94.96 &55.50 &75.23\\ 
ECPE-MLL &94.68 &48.48 &71.59\\
RankCP &97.30 &33.00 &65.15\\
\midrule
\multicolumn{4}{c}{\textbf{CEE Methods}}  \\
RoBERTa-Base &88.74 &64.28 &76.51\\ 
RoBERTa-Large &87.89 &66.23 &77.06\\ 
KEC &88.85 &66.55 &77.70\\ 
\midrule
KBCIN (Ours) &89.65 &\textbf{68.59} &\textbf{79.12}\\
\bottomrule& 
\end{tabular}
\caption{Comparison of our model against state-of-the-art baselines of CEE, ECE and ECPE.}
\label{tab2}
\end{table}

For CEE baselines, KBCIN outperforms the strong RoBERTa-Base even RoBERTa-Large, which just concatenate the target and candidate utterances with the dialogue history. And two instructive conclusions can be derived. On the one hand, it is of great importance to explicitly perform conversational context modeling and capture inter-utterance dependencies. On the other hand, accurate target-aware emotion cause reasoning also matters and causal clues provided by commonsense knowledge sever as the bridge to fill the reasoning gap. Although CSK is also utilised in KEC, it only focuses on the emotional level to reason the target emotion and the reasoning process is restricted within a certain range by a window, which may result in the problem of position bias and weaken the ability to generalize well on other datasets, where causal utterances are not located near the target utterance. Instead of depending on the position bias to identify causal utterances, KBCIN can fully understand the conversational context to enhance the deeper inter-utterance dependencies and measures all the candidate utterances through E-bridge and A-bridge to accurately reason the cause of the target emotion.

\subsection{Ablation Study}
We conduct ablation studies to verify the effectiveness of different modules proposed in our model.

\begin{table}
\footnotesize
\centering
\begin{tabular}{cccccc}
\toprule
\textbf{S-bridge} & \textbf{E-bridge} & \textbf{A-bridge} & \textbf{Pos. F1} & \textbf{macro F1} \\
\midrule
\usym{2714} &\usym{2714} &\usym{2714} &\textbf{68.59} &\textbf{79.12} \\
\usym{2718} &\usym{2714} &\usym{2714} &67.47 &78.63 \\
\usym{2714} &\usym{2718} &\usym{2714} &66.33 &77.78\\
\usym{2714} &\usym{2714} &\usym{2718} &66.92 &78.37\\
\usym{2718} &\usym{2718} &\usym{2718} &57.59 &71.81\\
\bottomrule
\end{tabular}
\caption{Results of ablation study.}
\label{tab3}
\end{table}

\subsubsection{Effect of Knowledge Bridge}

To investigate the impact of three bridges constructed by commonsense knowledge, we remove each one of the bridge individually. First, when removing the S-bridge, the way of conversational context modeling degrades to the interactions between pure context with attention mechanism. Dropped results in the second row of Table \ref{tab3} demonstrate the effectiveness of incorporating the S-bridge to enhance inter-utterance dependencies, leading to a comprehensive understanding of the conversational context. Also, by removing either the E-bridge or the A-bridge in the process of reasoning the target emotion, the performance of the model degrades to a certain extent. This suggests that both E-bridge and A-bridge play an important role in offering explicit causal clues to fill the reasoning gap between the target and candidate utterances. Further, the performance drops dramatically when we remove all three bridges, which verifies our analysis of the challenge of CEE that it requires the model to effectively capture causal clues contained in the context and accurately reason candidate utterances to the target emotion.

\begin{table}
\footnotesize
\centering
\begin{tabular}{ccc}
\toprule
\textbf{Emotion} & \textbf{Pos. F1} & \textbf{macro F1} \\
\midrule
Gold Emotion &\textbf{68.59} &\textbf{79.12} \\
Predicted Emotion &67.51 &78.43 \\
No Emotion &64.05 &76.73\\
\bottomrule
\end{tabular}
\caption{Results of our proposed KBCIN with different ways of using emotion information.}
\label{tab4}
\end{table}

\subsection{Impact of Emotion Information}
To further investigate the effect of emotion information of each utterance in the dialogue history, we either remove the emotion information or replace the gold emotion labels with the labels predicted by an emotion recognition model. Results are shown in Table \ref{tab4}. When removing the emotion information, the clear dropped performance is demonstrated in the last row of Table \ref{tab4}. It verifies the effectiveness of the emotion information to directly manifest the intra and inter emotional dependencies of speakers. And the reason why we test the performance of KBCIN with the predicted labels is that emotion recognition is the prerequisite process of emotion cause extraction under the circumstance of real applications, which means such gold emotion labels of utterances in the dialogue history may not be available in a practical emotion cause extraction system. To achieve this, we utilise an advanced emotion recognition model CauAIN \cite{cauain}. We train the model on the train set of DailyDialog \cite{dd} dataset and use it to predict the emotion labels of each utterance in the dialogue history from the test set of RECCON-DD. And the accuracy of the predicted labels recognized by CauAIN is 76.45. As shown in the second row in Table \ref{tab4}, the performance of KBCIN drops to a certain degree, but it still gains an obvious improvement compared to the model without any emotion information. On the one hand, 
\begin{figure}[htbp]
\centering
\includegraphics[width=0.4\textwidth]{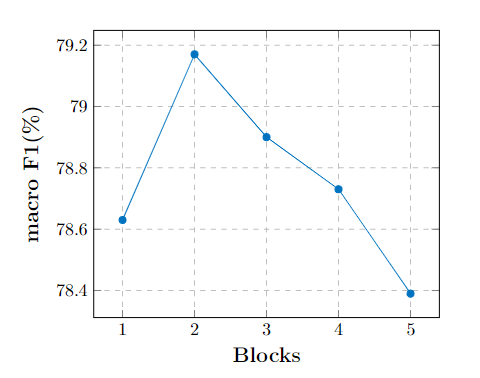}
\caption{Results of our proposed model with different numbers of Knowledge-Bridged Causal Interaction block.}
\label{curve}
\end{figure}
it further proves that even in the presence of false 
labels, emotion information of each candidate utterance 
is still very helpful in accurately identifying the cause of the target emotion. On the other hand, for the real application, it reminds us to jointly perform emotion recognition and emotion cause extraction in conversations, which could share the correlated emotion information between two tasks and alleviate the problem of error propagation resulting from using the emotion information in the two-stage way.

\subsection{Number of KBCI Blocks}
Since KBCI is the innovative and critical component of our model for effective conversational context modeling and accurate emotion cause reasoning, we adjust different numbers of KBCI blocks for a deeper analysis of the performance. Results are shown in Figure \ref{tab4}. With the increasing number of KBCI blocks from the range of 1 to 5, the model with two KBCI blocks achieves the best performance. On the one hand, the model with a single KBCI block may not comprehend deep inter-utterance dependencies contained in the conversational context and it is insufficient to effectively capture causal clues for the accurate reasoning between candidate utterances and the target. On the other hand, much more redundant causal information may be captured by the model with large numbers of block, which would weaken the performance of our model.

\begin{figure}[htbp]
\centering
\includegraphics[width=0.4\textwidth]{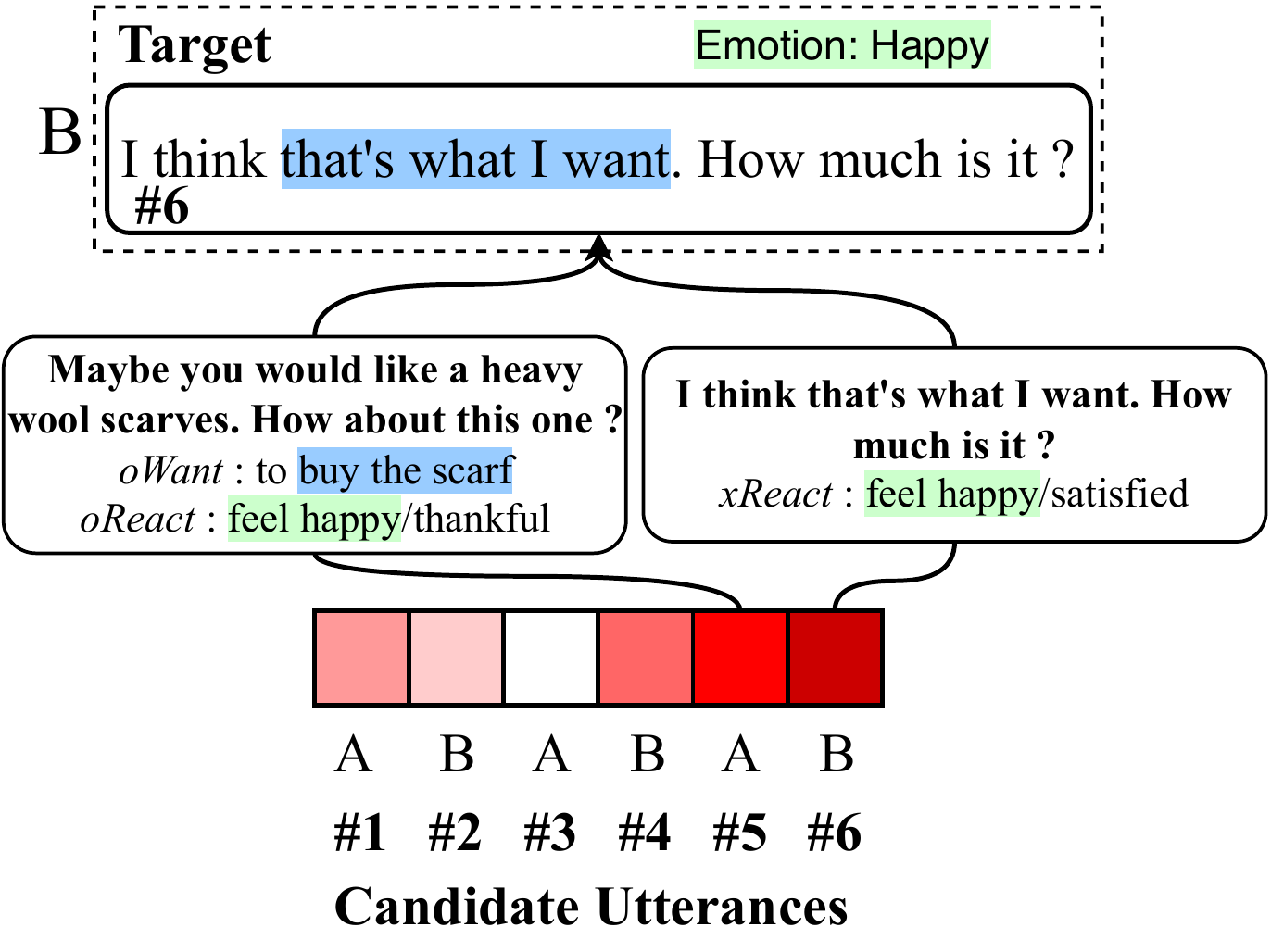}
\caption{A case that our model makes the right prediction, along with the visualizations for attention weights of the candidate utterances (obtained by summing up the measuring results after the process of Emotional Interaction and Actional Interaction from each block). The darker color mean larger attention weights.}
\label{case_study}
\end{figure}

\subsection{Case Study}
In Figure \ref{case_study}, we exemplify a conversation between two speakers A and B to demonstrate that our proposed KBCIN correctly identifies the casual utterances $\#5$ and $\#6$ for the target utterance $\#6$ with emotion \emph{happy}. The situation is that PersonB chooses suitable scarves in the store with the help of PersonA. According to the results of Emotional Interaction and Actional Interaction to reason the cause of the target emotion, we visualize the attention weights to manifest different importance of candidate utterances. On the one hand, the target utterance $\#6$ itself is the most attended causal utterance, where PersonA gets the exact scarf and the target emotion \emph{happy} could be implied by E-bridge. On the other hand, for another causal utterance $\#5$ that attends most by our model, PersonA recommends the heavy wool scarves for PersonB in the causal utterance $\#5$, which meets the demand of PersonA and results in the action tendency for PersonA to \emph{buy the scarf}. Also, PersonB would \emph{feel happy} because of the accurate recommendation to offer a pleasant shopping experience. Here, the reasoning process is aided by both E-bridge and A-bridge to capture potential causal clues for the right prediction.

\section{Conclusion}
In this paper, we propose novel Knowledge-Bridged Causal Interaction Network (KBCIN) for causal emotion entailment. Commonsense knowledge (CSK) is leveraged as three bridges to carry out effective conversational context modeling and accurate emotion cause reasoning. To be more specific, we abstract the conversation as a conversation graph and leverage the event-centered CSK as the semantics-level bridge (S-bridge) to enhance the deep inter-utterance dependencies by message passing on the graph with CSK-enhanced Graph Attention module. And the social-interaction CSK serves as emotion-level bridge (E-bridge) and action-level bridge (A-bridge) to provide explicit causal clues from the perspective of human's feeling and action tendency for Emotional Interaction module and Actional Interaction module, which fills the reasoning gap between candidate utterances and the target one. Experimental results on the benchmark dataset demonstrate the effectiveness of our proposed KBCIN.

For future work, we would explore how to combine the knowledge from pretrained language models with external knowledge base. Further, to improve the performance of the model under the circumstance of real applications, joint emotion cause recognition and emotion recognition are supposed to be explored to share the interchanged emotional information and alleviate the problem of error propagation.

\section*{Acknowledgments}
We thank the anonymous reviewers for their insightful comments and suggestions. This work was supported by the National Key RD Program of China via grant  2021YFF0901602 and the National Natural Science Foundation of China (NSFC) via grant 62176078.

\appendix

\label{sec:reference_examples}

\bibliography{aaai23}

\begin{thebibliography}{27}
\providecommand{\natexlab}[1]{#1}

\bibitem[{Bao et~al.(2022)Bao, Ma, Wei, Zhou, and Hu}]{ece-2022}
Bao, Y.; Ma, Q.; Wei, L.; Zhou, W.; and Hu, S. 2022.
\newblock Multi-Granularity Semantic Aware Graph Model for Reducing Position
  Bias in Emotion Cause Pair Extraction.
\newblock In \emph{Findings of the Association for Computational Linguistics:
  ACL 2022}, 1203--1213. Dublin, Ireland: Association for Computational
  Linguistics.

\bibitem[{Bosselut et~al.(2019)Bosselut, Rashkin, Sap, Malaviya, Celikyilmaz,
  and Choi}]{comet}
Bosselut, A.; Rashkin, H.; Sap, M.; Malaviya, C.; Celikyilmaz, A.; and Choi, Y.
  2019.
\newblock {COMET:} Commonsense Transformers for Automatic Knowledge Graph
  Construction.
\newblock In Korhonen, A.; Traum, D.~R.; and M{\`{a}}rquez, L., eds.,
  \emph{Proceedings of the 57th Conference of the Association for Computational
  Linguistics, {ACL} 2019, Florence, Italy, July 28- August 2, 2019, Volume 1:
  Long Papers}, 4762--4779. Association for Computational Linguistics.

\bibitem[{Ding et~al.(2019)Ding, He, Zhang, and Xia}]{ding2019}
Ding, Z.; He, H.; Zhang, M.; and Xia, R. 2019.
\newblock From Independent Prediction to Reordered Prediction: Integrating
  Relative Position and Global Label Information to Emotion Cause
  Identification.
\newblock In \emph{The Thirty-Third {AAAI} Conference on Artificial
  Intelligence, {AAAI} 2019, The Thirty-First Innovative Applications of
  Artificial Intelligence Conference, {IAAI} 2019, The Ninth {AAAI} Symposium
  on Educational Advances in Artificial Intelligence, {EAAI} 2019, Honolulu,
  Hawaii, USA, January 27 - February 1, 2019}, 6343--6350. {AAAI} Press.

\bibitem[{Ding, Xia, and Yu(2020{\natexlab{a}})}]{ecpe-2d}
Ding, Z.; Xia, R.; and Yu, J. 2020{\natexlab{a}}.
\newblock {ECPE}-2{D}: Emotion-Cause Pair Extraction based on Joint
  Two-Dimensional Representation, Interaction and Prediction.
\newblock In \emph{Proceedings of the 58th Annual Meeting of the Association
  for Computational Linguistics}, 3161--3170. Online: Association for
  Computational Linguistics.

\bibitem[{Ding, Xia, and Yu(2020{\natexlab{b}})}]{ecpe-mll}
Ding, Z.; Xia, R.; and Yu, J. 2020{\natexlab{b}}.
\newblock End-to-End Emotion-Cause Pair Extraction based on Sliding Window
  Multi-Label Learning.
\newblock In \emph{Proceedings of the 2020 Conference on Empirical Methods in
  Natural Language Processing (EMNLP)}, 3574--3583. Online: Association for
  Computational Linguistics.

\bibitem[{Gao et~al.(2021)Gao, Liu, Deng, Wang, Cao, Du, and Xu}]{empathy1}
Gao, J.; Liu, Y.; Deng, H.; Wang, W.; Cao, Y.; Du, J.; and Xu, R. 2021.
\newblock Improving Empathetic Response Generation by Recognizing Emotion Cause
  in Conversations.
\newblock In \emph{Findings of the Association for Computational Linguistics:
  EMNLP 2021}, 807--819. Punta Cana, Dominican Republic: Association for
  Computational Linguistics.

\bibitem[{Ghosal et~al.(2020)Ghosal, Majumder, Gelbukh, Mihalcea, and
  Poria}]{cosmic}
Ghosal, D.; Majumder, N.; Gelbukh, A.; Mihalcea, R.; and Poria, S. 2020.
\newblock {COSMIC}: {CO}mmon{S}ense knowledge for e{M}otion Identification in
  Conversations.
\newblock In \emph{Findings of the Association for Computational Linguistics:
  EMNLP 2020}, 2470--2481. Online: Association for Computational Linguistics.

\bibitem[{Gui et~al.(2016)Gui, Wu, Xu, Lu, and Zhou}]{ece-data}
Gui, L.; Wu, D.; Xu, R.; Lu, Q.; and Zhou, Y. 2016.
\newblock Event-Driven Emotion Cause Extraction with Corpus Construction.
\newblock In \emph{Proceedings of the 2016 Conference on Empirical Methods in
  Natural Language Processing}, 1639--1649. Austin, Texas: Association for
  Computational Linguistics.

\bibitem[{Hwang et~al.(2021)Hwang, Bhagavatula, Bras, Da, Sakaguchi, Bosselut,
  and Choi}]{csk}
Hwang, J.~D.; Bhagavatula, C.; Bras, R.~L.; Da, J.; Sakaguchi, K.; Bosselut,
  A.; and Choi, Y. 2021.
\newblock (Comet-) Atomic 2020: On Symbolic and Neural Commonsense Knowledge
  Graphs.
\newblock In \emph{Thirty-Fifth {AAAI} Conference on Artificial Intelligence,
  {AAAI} 2021, Thirty-Third Conference on Innovative Applications of Artificial
  Intelligence, {IAAI} 2021, The Eleventh Symposium on Educational Advances in
  Artificial Intelligence, {EAAI} 2021, Virtual Event, February 2-9, 2021},
  6384--6392. {AAAI} Press.

\bibitem[{Lewis et~al.(2020)Lewis, Liu, Goyal, Ghazvininejad, Mohamed, Levy,
  Stoyanov, and Zettlemoyer}]{bart}
Lewis, M.; Liu, Y.; Goyal, N.; Ghazvininejad, M.; Mohamed, A.; Levy, O.;
  Stoyanov, V.; and Zettlemoyer, L. 2020.
\newblock {BART:} Denoising Sequence-to-Sequence Pre-training for Natural
  Language Generation, Translation, and Comprehension.
\newblock In Jurafsky, D.; Chai, J.; Schluter, N.; and Tetreault, J.~R., eds.,
  \emph{Proceedings of the 58th Annual Meeting of the Association for
  Computational Linguistics, {ACL} 2020, Online, July 5-10, 2020}, 7871--7880.
  Association for Computational Linguistics.

\bibitem[{Li et~al.(2022)Li, Meng, Lin, Liu, Fu, Cao, Wang, and Zhou}]{kec}
Li, J.; Meng, F.; Lin, Z.; Liu, R.; Fu, P.; Cao, Y.; Wang, W.; and Zhou, J.
  2022.
\newblock Neutral Utterances are Also Causes: Enhancing Conversational Causal
  Emotion Entailment with Social Commonsense Knowledge.
\newblock In Raedt, L.~D., ed., \emph{Proceedings of the Thirty-First
  International Joint Conference on Artificial Intelligence, {IJCAI} 2022,
  Vienna, Austria, 23-29 July 2022}, 4209--4215. ijcai.org.

\bibitem[{Li et~al.(2019)Li, Feng, Wang, and Zhang}]{li2019}
Li, X.; Feng, S.; Wang, D.; and Zhang, Y. 2019.
\newblock Context-aware emotion cause analysis with multi-attention-based
  neural network.
\newblock \emph{Knowl. Based Syst.}, 174: 205--218.

\bibitem[{Li et~al.(2017)Li, Su, Shen, Li, Cao, and Niu}]{dd}
Li, Y.; Su, H.; Shen, X.; Li, W.; Cao, Z.; and Niu, S. 2017.
\newblock DailyDialog: {A} Manually Labelled Multi-turn Dialogue Dataset.
\newblock In Kondrak, G.; and Watanabe, T., eds., \emph{Proceedings of the
  Eighth International Joint Conference on Natural Language Processing,
  {IJCNLP} 2017, Taipei, Taiwan, November 27 - December 1, 2017 - Volume 1:
  Long Papers}, 986--995. Asian Federation of Natural Language Processing.

\bibitem[{Liu et~al.(2019)Liu, Ott, Goyal, Du, Joshi, Chen, Levy, Lewis,
  Zettlemoyer, and Stoyanov}]{roberta}
Liu, Y.; Ott, M.; Goyal, N.; Du, J.; Joshi, M.; Chen, D.; Levy, O.; Lewis, M.;
  Zettlemoyer, L.; and Stoyanov, V. 2019.
\newblock RoBERTa: {A} Robustly Optimized {BERT} Pretraining Approach.
\newblock \emph{CoRR}.

\bibitem[{Majumder et~al.(2019)Majumder, Poria, Hazarika, Mihalcea, Gelbukh,
  and Cambria}]{dialoguernn}
Majumder, N.; Poria, S.; Hazarika, D.; Mihalcea, R.; Gelbukh, A.; and Cambria,
  E. 2019.
\newblock Dialoguernn: An attentive rnn for emotion detection in conversations.
\newblock In \emph{Proceedings of the AAAI conference on artificial
  intelligence}, volume~33, 6818--6825.

\bibitem[{Moors and A.(2013)}]{theory}
Moors; and A. 2013.
\newblock On the causal role of appraisal in emotion.
\newblock \emph{Emotion Review Journal of the International Society for
  Research on Emotions}, 5(2): 132--140.

\bibitem[{Peng et~al.(2022)Peng, Hu, Xing, Xie, Sun, and Li}]{es_peng}
Peng, W.; Hu, Y.; Xing, L.; Xie, Y.; Sun, Y.; and Li, Y. 2022.
\newblock Control Globally, Understand Locally: {A} Global-to-Local
  Hierarchical Graph Network for Emotional Support Conversation.
\newblock In Raedt, L.~D., ed., \emph{Proceedings of the Thirty-First
  International Joint Conference on Artificial Intelligence, {IJCAI} 2022,
  Vienna, Austria, 23-29 July 2022}, 4324--4330. ijcai.org.

\bibitem[{Poria et~al.(2021)Poria, Majumder, Hazarika, Ghosal, Bhardwaj, Jian,
  Hong, Ghosh, Roy, Chhaya, Gelbukh, and Mihalcea}]{reccon}
Poria, S.; Majumder, N.; Hazarika, D.; Ghosal, D.; Bhardwaj, R.; Jian, S.
  Y.~B.; Hong, P.; Ghosh, R.; Roy, A.; Chhaya, N.; Gelbukh, A.~F.; and
  Mihalcea, R. 2021.
\newblock Recognizing Emotion Cause in Conversations.
\newblock \emph{Cogn. Comput.}, 13(5): 1317--1332.

\bibitem[{Shen et~al.(2021)Shen, Wu, Yang, and Quan}]{dag-erc}
Shen, W.; Wu, S.; Yang, Y.; and Quan, X. 2021.
\newblock Directed Acyclic Graph Network for Conversational Emotion
  Recognition.
\newblock In \emph{Proceedings of the 59th Annual Meeting of the Association
  for Computational Linguistics and the 11th International Joint Conference on
  Natural Language Processing (Volume 1: Long Papers)}, 1551--1560. Online:
  Association for Computational Linguistics.

\bibitem[{Turcan et~al.(2021)Turcan, Wang, Anubhai, Bhattacharjee,
  Al{-}Onaizan, and Muresan}]{adapted}
Turcan, E.; Wang, S.; Anubhai, R.; Bhattacharjee, K.; Al{-}Onaizan, Y.; and
  Muresan, S. 2021.
\newblock Multi-Task Learning and Adapted Knowledge Models for Emotion-Cause
  Extraction.
\newblock In Zong, C.; Xia, F.; Li, W.; and Navigli, R., eds., \emph{Findings
  of the Association for Computational Linguistics: {ACL/IJCNLP} 2021, Online
  Event, August 1-6, 2021}, volume {ACL/IJCNLP} 2021 of \emph{Findings of
  {ACL}}, 3975--3989. Association for Computational Linguistics.

\bibitem[{Vaswani et~al.(2017)Vaswani, Shazeer, Parmar, Uszkoreit, Jones,
  Gomez, Kaiser, and Polosukhin}]{trans}
Vaswani, A.; Shazeer, N.; Parmar, N.; Uszkoreit, J.; Jones, L.; Gomez, A.~N.;
  Kaiser, L.; and Polosukhin, I. 2017.
\newblock Attention is All you Need.
\newblock In Guyon, I.; von Luxburg, U.; Bengio, S.; Wallach, H.~M.; Fergus,
  R.; Vishwanathan, S. V.~N.; and Garnett, R., eds., \emph{Advances in Neural
  Information Processing Systems 30: Annual Conference on Neural Information
  Processing Systems 2017, December 4-9, 2017, Long Beach, CA, {USA}},
  5998--6008.

\bibitem[{Velickovic et~al.(2018)Velickovic, Cucurull, Casanova, Romero,
  Li{\`{o}}, and Bengio}]{gat}
Velickovic, P.; Cucurull, G.; Casanova, A.; Romero, A.; Li{\`{o}}, P.; and
  Bengio, Y. 2018.
\newblock Graph Attention Networks.
\newblock In \emph{6th International Conference on Learning Representations,
  {ICLR} 2018, Vancouver, BC, Canada, April 30 - May 3, 2018, Conference Track
  Proceedings}. OpenReview.net.

\bibitem[{Wei, Zhao, and Mao(2020)}]{rankcp}
Wei, P.; Zhao, J.; and Mao, W. 2020.
\newblock Effective Inter-Clause Modeling for End-to-End Emotion-Cause Pair
  Extraction.
\newblock In \emph{Proceedings of the 58th Annual Meeting of the Association
  for Computational Linguistics}, 3171--3181. Online: Association for
  Computational Linguistics.

\bibitem[{Xia and Ding(2019)}]{ecpe}
Xia, R.; and Ding, Z. 2019.
\newblock Emotion-Cause Pair Extraction: A New Task to Emotion Analysis in
  Texts.
\newblock In \emph{Proceedings of the 57th Annual Meeting of the Association
  for Computational Linguistics}, 1003--1012. Florence, Italy: Association for
  Computational Linguistics.

\bibitem[{Xia, Zhang, and Ding(2019)}]{xia2019}
Xia, R.; Zhang, M.; and Ding, Z. 2019.
\newblock {RTHN:} {A} RNN-Transformer Hierarchical Network for Emotion Cause
  Extraction.
\newblock In Kraus, S., ed., \emph{Proceedings of the Twenty-Eighth
  International Joint Conference on Artificial Intelligence, {IJCAI} 2019,
  Macao, China, August 10-16, 2019}, 5285--5291. ijcai.org.

\bibitem[{Yan et~al.(2021)Yan, Gui, Pergola, and He}]{kag}
Yan, H.; Gui, L.; Pergola, G.; and He, Y. 2021.
\newblock Position Bias Mitigation: {A} Knowledge-Aware Graph Model for Emotion
  Cause Extraction.
\newblock In Zong, C.; Xia, F.; Li, W.; and Navigli, R., eds.,
  \emph{Proceedings of the 59th Annual Meeting of the Association for
  Computational Linguistics and the 11th International Joint Conference on
  Natural Language Processing, {ACL/IJCNLP} 2021, (Volume 1: Long Papers),
  Virtual Event, August 1-6, 2021}, 3364--3375. Association for Computational
  Linguistics.

\bibitem[{Zhao, Zhao, and Lu(2022)}]{cauain}
Zhao, W.; Zhao, Y.; and Lu, X. 2022.
\newblock CauAIN: Causal Aware Interaction Network for Emotion Recognition in
  Conversations.
\newblock In Raedt, L.~D., ed., \emph{Proceedings of the Thirty-First
  International Joint Conference on Artificial Intelligence, {IJCAI} 2022,
  Vienna, Austria, 23-29 July 2022}, 4524--4530. ijcai.org.

\end{thebibliography}

\end{document}